\documentclass[10pt,twocolumn,letterpaper]{article}

\usepackage[pagenumbers]{cvpr} 

\usepackage{graphicx}
\usepackage{amsmath}
\usepackage{amssymb}
\usepackage{booktabs}
\usepackage{colortbl}
\usepackage{multirow}
\usepackage{pifont}
\usepackage{subcaption}
\usepackage{enumitem}
\usepackage{colortbl}
\usepackage{soul}
\usepackage{color}
\usepackage{bm}
\usepackage{lipsum} 
\usepackage{footnote}
\usepackage[symbol]{footmisc}
\usepackage{setspace}
\usepackage{longtable}

\usepackage{flushend}

\usepackage{listings}
\definecolor{apricot}{rgb}{0.98, 0.81, 0.69}
\definecolor{ao}{rgb}{0.0, 0.0, 1.0}
\definecolor{bleudefrance}{rgb}{0.19, 0.55, 0.91}
\definecolor{brandeisblue}{rgb}{0.0, 0.44, 1.0}
\definecolor{yt}{rgb}{0.97, 0.97, 1.0}
\DeclareRobustCommand{\hlorange}[1]{{\sethlcolor{apricot}\hl{#1}}}
\DeclareRobustCommand{\hlblue}[1]{{\sethlcolor{brandeisblue}\hl{#1}}}

\usepackage[pagebackref=true,breaklinks=true,colorlinks,bookmarks=false]{hyperref}
\hypersetup{
	plainpages=false,
	colorlinks=true,              %
	linkcolor=blue,               %
	anchorcolor=blue,             %
	citecolor=blue,               %
	filecolor=blue,               %
	urlcolor=blue,                %
	pdfview=FitH,                 %
	pdfstartview=FitH,            %
	pdfpagelayout=SinglePage      %
}

\usepackage[capitalize]{cleveref}
\crefname{section}{Sec.}{Secs.}
\Crefname{section}{Section}{Sections}
\Crefname{table}{Table}{Tables}
\crefname{table}{Tab.}{Tabs.}

\definecolor{mygray}{gray}{0.6}

\newcommand{\tablestyle}[2]{\setlength{\tabcolsep}{#1}\renewcommand{\arraystretch}{#2}\centering\footnotesize}
\newcommand{\improve}[1]{\textcolor[rgb]{0.22,0.463,0.114}{\scriptsize{#1}}}

\def\cvprPaperID{413} 
\def\confName{CVPR}
\def\confYear{2023}

\begin{document}

\title{Class Prototypes based Contrastive Learning for Classifying Multi-Label and Fine-Grained Educational Videos}

\author{Rohit Gupta~\textsuperscript{1} \footnotemark
\and
Anirban Roy~\textsuperscript{2}
\and 
Claire Christensen~\textsuperscript{2}
\and 
Sujeong Kim~\textsuperscript{2}
\and 
Sarah Gerard~\textsuperscript{2}
\and 
Madeline Cincebeaux~\textsuperscript{2}
\and
Ajay Divakaran~\textsuperscript{2}
\and 
Todd Grindal~\textsuperscript{2}
\and
Mubarak Shah~\textsuperscript{1}
\and
{\textsuperscript{1}~\small Center for Research in Computer Vision, University of Central Florida} \\
{\tt\small rohitg@knights.ucf.edu, shah@crcv.ucf.edu}
\and
{\textsuperscript{2}~\small SRI International} \\
{\tt\small anirban.roy@sri.com}
}
\maketitle
\footnotetext[1]{Work partly done during an internship at SRI International.}

\begin{abstract}
The recent growth in the consumption of online media by children during early childhood necessitates data-driven tools enabling educators to filter out appropriate educational content for young learners. This paper presents an approach for detecting educational content in online videos.  We focus on two widely used educational content classes: literacy and math. For each class, we choose prominent codes (sub-classes) based on the Common Core Standards. For example, literacy codes include `letter names', `letter sounds', and math codes include `counting', `sorting'. We pose this as a fine-grained multilabel classification problem as videos can contain multiple types of educational content and the content classes can get visually similar (e.g., `letter names' vs `letter sounds'). We propose a novel class prototypes based supervised contrastive learning approach that can handle fine-grained samples associated with multiple labels. We learn a class prototype for each class and a loss function is employed to minimize the distances between a class prototype and the samples from the class. Similarly, distances between a class prototype and the samples from other classes are maximized. As the alignment between visual and audio cues are crucial for effective comprehension, we consider a multimodal transformer network to capture the interaction between visual and audio cues in videos while learning the embedding for videos. For evaluation, we present a dataset, APPROVE, employing educational videos from YouTube labeled with fine-grained education classes by education researchers. APPROVE consists of 193 hours of expert-annotated videos with 19 classes. The proposed approach outperforms strong baselines on APPROVE and other benchmarks such as Youtube-8M, and COIN. The dataset is available at \url{https://nusci.csl.sri.com/project/APPROVE}.

\end{abstract}

\section{Introduction}
\label{sec:intro}

\vspace{-2mm}
With the expansion of internet access and the ubiquitous availability of smart devices, children increasingly spend a significant amount of time watching online videos. A recent nationally representative survey reported that 89\% of parents of children aged 11 or younger say their child watches videos on YouTube~\cite{Auxier2020}. Moreover, it is estimated that young children in the age range of two to four years consume 2.5 hours and five to eight years consume 3.0 hours per day on average \cite{Rideout2019,screentime1}. Childhood is typically a key period for education, especially for learning basic skills such as literacy and math~\cite{hemphill2008importance, jordan2009early}. Unlike generic online videos, watching appropriate educational videos supports healthy child development and learning~\cite{hurwitz2019getting,hurwitz2020raising,burkhardt2022meta}. Thus, analyzing the content of these videos may help parents, teachers, and media developers increase young children’s exposure to high-quality education videos, which has been shown to produce meaningful learning gains~\cite{hurwitz2019getting}. As the amount of online content produced grows exponentially, automated content understanding methods are essential to facilitate this.


In this work, given a video, our goal is to determine whether the video contains any educational content and characterize the content. 
Detecting educational content requires identifying multiple distinct types of content in a video while distinguishing between similar content types. The task is challenging as the education codes by Common Core Standards \cite{national2010common,porter2011common} can be similar such as `letter names' and `letter sounds', where the former focuses on the name of the letter and the latter is based on the phonetic sound of the letter. Also,  understanding education content requires analyzing both visual and audio cues simultaneously as both signals are to be present to ensure effective learning \cite{national2010common,porter2011common}. This is in contrast to  standard video classification benchmarks such as the sports or generic YouTube videos in UCF101~\cite{soomro2012ucf101} Kinetics400~\cite{k700}, YouTube-8M~\cite{abu2016youtube}, where visual cues are often sufficient to detect the different classes. Finally, unlike standard well-known action videos, education codes are more structured and not accessible to common users. Thus, it requires a carefully curated set of videos and expert annotations to create a dataset to enable a data-driven approach. In this work, we focus on two widely used educational content classes: literacy and math. For each class, we choose prominent codes (sub-classes) based on the Common Core Standards that outline age-appropriate learning standards~\cite{national2010common,porter2011common}. For example, literacy codes include `letter names', `letter sounds', `rhyming', and math codes include `counting', `addition subtraction', `sorting', `analyze shapes'.



We formulate the problem as a multilabel fine-grained video classification task as a video may contain multiple types of content that can be similar. We employ multimodal cues since besides visual cues, audio cues provide important cues to distinguish between similar types of educational content.  We propose a class prototypes based supervised contrastive learning approach to address the above-mentioned challenges. We learn a prototype embedding for each class. Then a loss function is employed to minimize the distance between a class prototype and the samples associated with the class label. Similarly, the distance between a class prototype and the samples without that class label is maximized. This is unlike the standard supervised contrastive learning setup where inter-class distance is maximized and intra-class distance is minimized by considering classwise positive and negative samples. This approach is shown to be effective for single-label setups~\cite{supcon}. However, it is not straightforward to extend this for the proposed multilabel setup as samples cannot be identified as positive or negative due to the multiple labels. We jointly learn the embedding of the class prototypes and the samples. The embeddings are learned by a multimodal transformer network (MTN) that captures the interaction between visual and audio cues in videos. We employ automatic speech recognition (ASR) to transcribe text from the audio. The MTN consists of video and text encoders that learn modality-specific embedding and a cross-attention mechanism is employed to capture the interaction between them. The MTN is end-to-end learned through the contrastive loss.

Due to the lack of suitable datasets for evaluating fine-grained classification of education videos, we propose a new dataset, called APPROVE, of curated YouTube videos annotated with educational content. We follow Common Core Standards \cite{national2010common,porter2011common} to select education content suitable for the kindergarten level. We consider two high-level classes of educational content: literacy and math. For each of these content classes, we select a set of codes. For the literacy class, we select 7 codes and for the math class, we select 11 codes. Each video is associated with multiple labels corresponding to these codes. The videos are annotated by trained education researchers following standard validation protocol \cite{Radesky2020} to ensure correctness. APPROVE also consists of carefully chosen background videos, i.e., without educational content, that are visually similar to the videos with educational content. APPROVE consists of 193 hours of expert-annotated videos with 19 classes (7 literacy codes, 11 math codes, and a background) where each video has 3 labels on average.

\noindent Our contributions can be summarized as follows:

\begin{itemize}[noitemsep,nolistsep]
    \item APPROVE, a fine-grained multi-label dataset of education videos, to promote exploration in this field.
    \item Class prototypes based contrastive learning framework along with a multi-modal fusion transformer suitable for the problem where videos have multiple fine-grained labels. 
    \item Outperforming relevant baselines on three datasets: APPROVE, YouTube-8M~\cite{abu2016youtube} and COIN~\cite{coin}.
\end{itemize}

\begin{figure*}[t]
    \centering
    \includegraphics[width=0.99\textwidth]{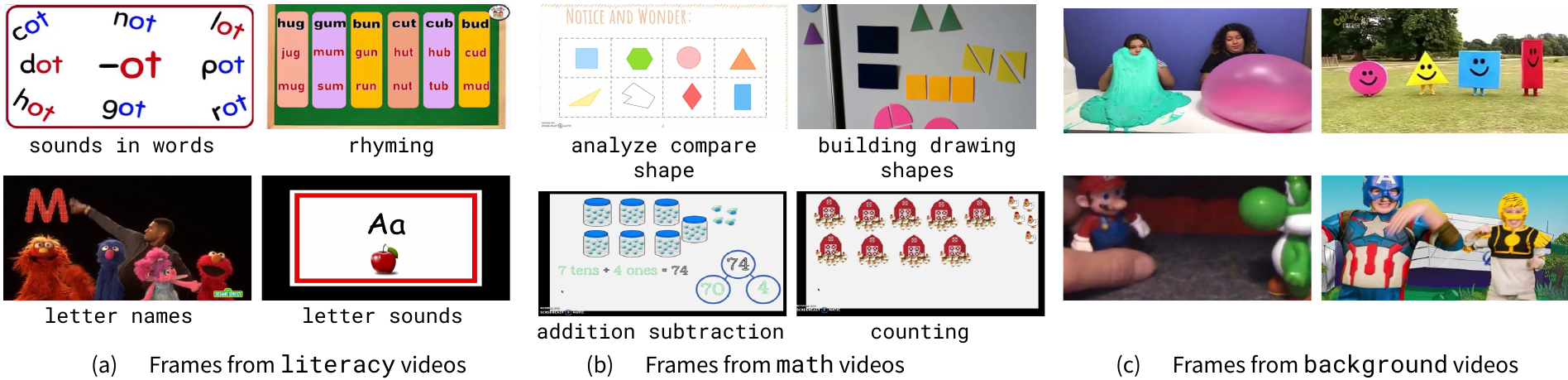}
    \vspace{-2mm}
    \caption{Sample video frames from the APPROVE dataset. Videos belong to the \textbf{(a) literacy classes},  \textbf{(b) math classes}, and  \textbf{(c) background}. Background videos do not contain educational content but share visual similarities with educational videos. The videos are labeled with fine-grained sub-classes, e.g., \texttt{letter names} vs \texttt{letter sounds}.}
    \label{fig:approve}
    \vspace{-6mm}
\end{figure*}
\section{Related Works}
\label{sec:rel_work}
\vspace{-1mm}


\noindent \textbf{Self-Supervised Contastive Learning (CL)}  has been an effective paradigm for visual representation learning. Methods such as SimCLR~\cite{simclr, simclrv2}, MoCo (Momentum Contrastive learning)~\cite{moco, mocov2, mocov3}, Augmented Multiscale Deep InfoMax (AMDIM)~\cite{amdim}, Contrastive Predictive Coding (CPC)~\cite{cpc}, Temporal Contrastive Learning (TCLR)~\cite{daveTCLR2022} and DLCL~\cite{9506759} have achieved strong performance on image and video classification benchmarks. The shared property between these CL frameworks is that data augmentation is used to generate positive pairs for CL from a single instance, where other data instances are treated as negatives. Prototypical Contrastive Learning (PCL)~\cite{li2021prototypical} extends self-supervised contrastive learning with the idea of clustering data representations during training to generate unsupervised prototypes which represent \textit{intra-class variation}. We utilize \textit{class prototypes} instead in the supervised setting, to learn fine-grained distinctions \textit{between classes}.

\noindent \textbf{Supervised CL} methods such as SupCon~\cite{supcon} utilize labels to enhance  contrastive learning by forming positive and negative pairs using labels instead of data augmentation. Supervised Contrastive Learning has also been used for other tasks such as image segmentation~\cite{conseg} and classification in the presence of noisy labels~\cite{connoise}. Hierarchical CL\cite{himulcon} extends SupCon to the hierarchical classification case. However, SupCon cannot be extended to the multi-label case in a straightforward manner, as pairs of data samples with multiple labels cannot be clearly classified just into positives and negatives.


\vspace{1mm}
\noindent \textbf{Weakly-Supervised Multi-Modal CL:} Weakly aligned text-image/video datasets scraped from the web such as Conceptual Captions~\cite{sharma-etal-2018-conceptual} and WebVid-10M~\cite{Bain21} enable learning of multi-modal representations. CLIP~\cite{radford2021learning} applies a cross-modal contrastive loss to train individual text and image encoders. Everything at Once~\cite{shvetsova2022everything} is able to additionally utilize the audio modality and incorporates a pairwise fusion encoder which encodes pairs of modalities, as a result, 6 forward passes of the fusion model are required for 3 modalities.  MASK~\cite{Swetha_2023_ICCV} proposes tri-modal alignment using a Sinkhorn based method for multi-attribute clustering, Frozen in Time~\cite{Bain21} is able to utilize both image-text and video-text datasets through the use of a Space-Time Transformer Visual Encoder. Visual Conditioned GPT~\cite{Luo2022} uses a single cross-attention fusion layer to combine pre-trained CLIP text and visual features. Flamingo~\cite{alayrac2022flamingo} adds cross-attention layers interleaved with language decoder layers to fuse visual information into text generation. MERLOT~\cite{zellers2021merlot, Zellers_2022_CVPR} and Triple Contrastive Learning~\cite{yang2022vision} combine contrastive learning and generative language modeling to learn aligned text-image representations.


\vspace{1mm}
\noindent \textbf{Supervised Multi-Modal Learning}: Supervised Multi-Modal Learning typically relies on crowd-captioned datasets such as Flickr30k~\cite{young-etal-2014-image} and MS-COCO Captions~\cite{chen2015microsoft}. Some prior works such as OSCAR~\cite{oscar} and VinVL~\cite{vinvl} have utilized pre-trained object detectors and multi-modal transformers to learn image captioning using supervised aligned datasets. BLIP~\cite{li2022blip} takes a hybrid approach where it bootstraps an image captioner using a labeled dataset and uses it to generate captions for web images. This generated corpus is then filtered and used for learning an aligned representation. ALign BEfore Fuse~\cite{ALBEF} highlights the importance of aligning text and image tokens before fusing them using a multi-modal transformer.

In this paper, we focus on the fine-grained classification of multilabel educational videos. Due to the lack of suitable datasets, we propose  a new dataset, APPROVE, which is described next.

\vspace{-2mm}
\section{APPROVE Dataset}
\label{sec:dataset}
\vspace{-2mm}

We propose a dataset, called APPROVE, of curated YouTube videos annotated with educational content. APPROVE consists of 193 hours of expert-annotated videos with 19 classes (7 literacy codes, 11 math, and background) and each video is associated with approximately 3 labels on average. We follow the Common Core Standards \cite{national2010common,porter2011common} to select education content suitable for kindergarten level. The Common Core Standards outline what students are expected to know and do at various age ranges and grades. This is a widely accepted standard followed by a range of educators.  We consider two high-level classes of educational content: literacy and math. For each of these content classes, we select a set of codes. For the literacy class, we select 7 codes including letter names, letter sounds, follow words, sight words, letters in words, sounds in words, and rhyming. For the math class, we select 11 codes including counting, individual number, comparing groups, addition subtraction, measurable attributes, sorting, spatial language, shape identification, building drawing, analyzing and comparing shapes. More details about the standard and the description of the codes are provided in the supplementary material. APPROVE also consists of carefully chosen background videos, i.e., without educational content, that are visually similar to the videos with educational content. We present frames corresponding to these classes in Fig.~\ref{fig:approve}.

\begin{figure}[ht]
    \centering
    \includegraphics[width=0.8\linewidth]{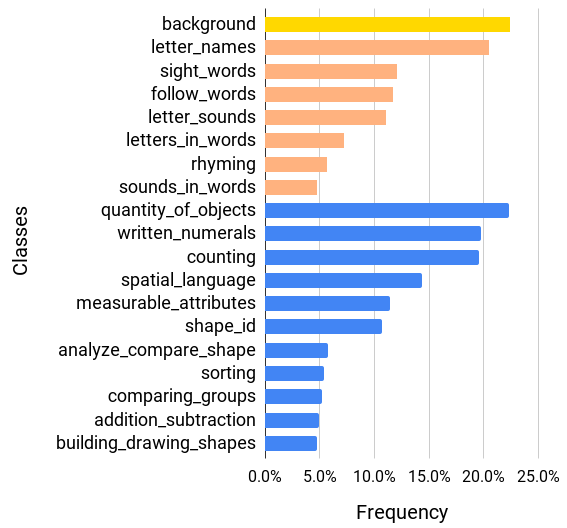}
    \vspace{-2mm}
    \caption{Frequency of the classes in APPROVE. \\ Math codes are in \textbf{\hlorange{Orange}} and literacy codes in \textbf{\textcolor{yt}{\hlblue{Blue}}}.}
    \label{fig:label-hours}
    \vspace{-5mm}
\end{figure}

\begin{figure}[ht]
    \centering
    \includegraphics[width=0.7\linewidth]{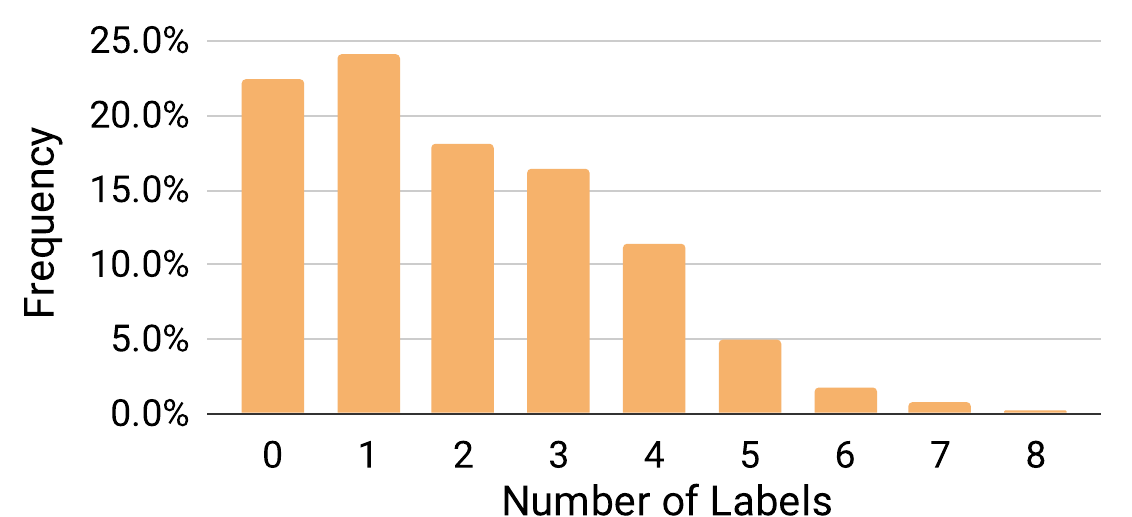}
    \vspace{-2mm}
    \caption{Distribution of the number of labels per video.}
    \label{fig:label-dist}
    \vspace{-3mm}
\end{figure}

To ensure the quality and correctness of the annotations, we consider educational researchers to annotate the videos and follow a standard validation protocol \cite{Radesky2020}. Each annotator is trained by an expert and annotations on a selected set are examined before engaging the annotator for the final annotation. Annotators start once they reach more than 90\% agreement with the expert. Further, we estimate inter-annotator consistency to filter out anomalies. Details about the validation process are provided in the supplementary material. It takes a month to train an education researcher to match expert-level coding accuracy. On average, it takes the trained annotators 1 min to annotate 1 min of video. 

The videos are curated from YouTube and are annotated by the trained annotators to determine educational content in them. Each video can have multiple class labels that are quite similar making the task a multi-label and fine-grained classification problem. For example `letter names' and `letter sounds' where visual letters are shown in both but in 'letter sounds', the phonetic sound on the letter is emphasized (Fig.~\ref{fig:approve} (a)). Similarly, in both `build and draw shapes' and `analyzing and comparing shapes', multiple shapes can appear but the latter focuses on comparing multiple shapes by shape and size (Fig.~\ref{fig:approve} (b)). Class-wise stats are presented in Fig.~\ref{fig:label-hours}. Note that the task is different from common video classification setups where either multi-label or fine-grained aspects are dealt separately. Single-label datasets such as HMDB51~\cite{kuehne2011hmdb}, UCF101~\cite{soomro2012ucf101}, Kinetics700~\cite{k700} and multi-label ones such as Charades~\cite{charades} are widely used benchmarks for this problem. YouTube-Birds and YouTube-Cars~\cite{zhu2018fine} are analogous datasets for object recognition from videos. Multi-Sports~\cite{li2021multisports} and FineGym~\cite{shao2020finegym} label fine-grained action classes for sports. VideoQA datasets~\cite{swetha2025implicitqa,swetha2025timelogic} test a broader range of visual skills, however rely on language based evaluation, making them unsuitable for evaluating pure vision representations. HVU~\cite{diba2020large} also adds scenes and attributes annotations along with action and objects. However, action, object and scene recognition are not enough for fine-grained video understanding. For instance, videos from a given education provider might share similar objects (person, chalkboard, etc.) and actions (writing on chalkboard) while covering different topics (counting, shape recognition etc.) in each video.
\begin{table}[ht]
\vspace{-2mm}
\small
\begin{tabular}{lccccc}
\toprule
Dataset & \begin{tabular}[x]{@{}c@{}} Size\\(in hr)\end{tabular} & Multi-Label & \begin{tabular}[x]{@{}c@{}}Fine\\Grained\end{tabular}  & Type & Annotators \\ 

\cmidrule(r){1-6}
\multicolumn{6}{l}{\textbf{Action Recognition}} \\
HMDB & 5 & \ding{56} & \ding{56} & V & Authors  \\ 
UCF  & 27 & \ding{56} & \ding{56} & V & Authors  \\
Kinetics & 800 &  \ding{56} & \ding{56} & V+A  & Crowd  \\ \cmidrule(r){1-6}
\multicolumn{6}{l}{\textbf{Video Classification}} \\
COIN & 476 & \ding{56} & \ding{56} & V+A & Crowd\\
YT-8M & - & \ding{52} & \ding{56} &  F  &  Machine \\ 
\cmidrule(r){1-6}
\textbf{APPROVE} & 193 & \ding{52} & \ding{52} & V+T+A & \textbf{Experts}  \\ \bottomrule
\end{tabular}
\caption{APPROVE dataset compared with selected prior datasets.\\ V$\rightarrow$Video Frames, A$\rightarrow$Audio, T$\rightarrow$Text, F$\rightarrow$Features only.}
\vspace{-3mm}
\end{table}

\section{Proposed Approach}
\label{sec:approach}
\vspace{-2mm}

In this section, we first describe the proposed class prototypes based contrastive learning framework suitable for videos containing multiple educational codes. Then we present the approach to learning the class prototypes and finally describe the multimodal transformer network that learns features by fusing visual and text cues from videos. 

\subsection{Class prototypes based contrastive learning}
\begin{figure*}[t]
    \centering
    \includegraphics[width=0.85\textwidth]{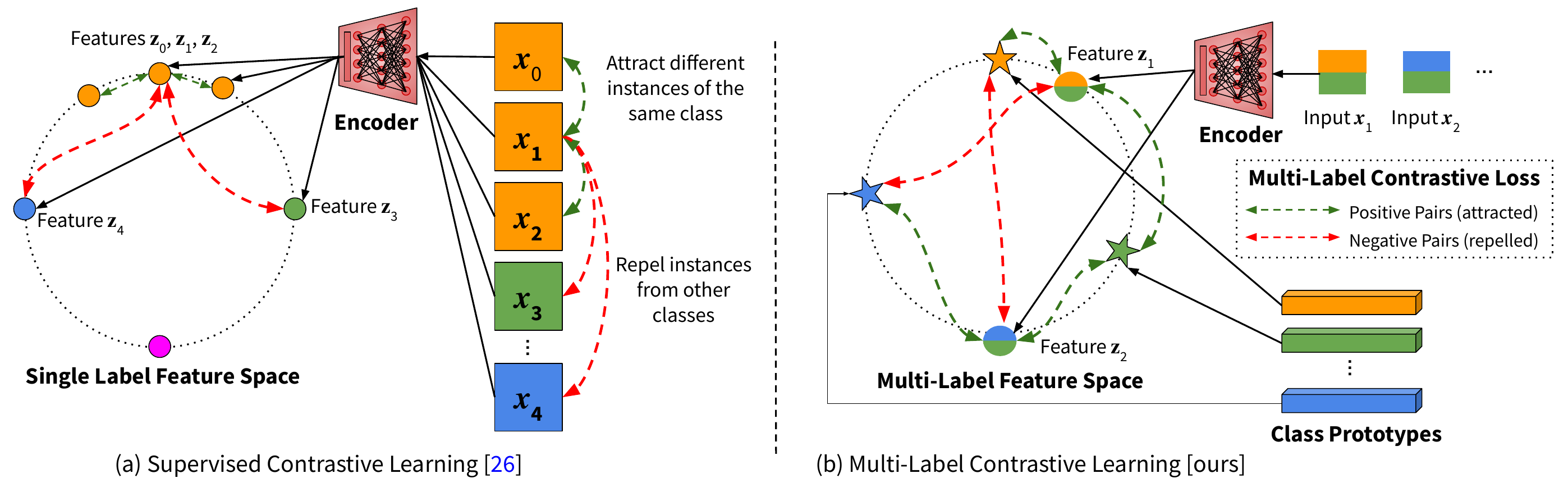}
    \vspace{-3mm}
    \caption{Contrastive learning operates on the feature space by bringing the representations of similar samples close and pushing distinct samples apart. Prior work in \textbf{(a) Supervised Contrastive Learning}~\cite{supcon} trains the network by treating instances from the same class as positive pairs and instances from different classes as negative pairs. This approach doesn't generalize to multi-label classification tasks, as some instance pairs have partially overlapping labels. We propose the use of class prototypes to enable \textbf{(b) Multi-Label Prototypes Contrastive Learning}. Each sample and the class prototypes corresponding to the labels associated with the sample are treated as positive pairs. Similarly, negative pairs are determined based on the missing class labels.  Prototypes are represented by stars ($\star$) and inputs as circles ($\circ$) colored with all their relevant labels. We discuss strategies for initializing and learning the label prototypes in Sec.~\ref{sec:prototypes}.}
    \label{fig:lcloss}
    \vspace{-2mm}
\end{figure*}

\vspace{-2mm}
\label{sec:loss}
In a contrastive learning framework, feature representations are typically learned by simultaneously minimizing the distance between positive samples and maximizing the distance between negative samples (See Figure \ref{fig:lcloss}.(a)). The positive and negative samples are determined with respect to an anchor sample usually based on the class labels. For example, supervised contrastive learning (SupCon)~\cite{supcon} learns a representation to minimize the intra-class distances and maximize inter-class distances. We denote $\bm{x}_i$ and $y_i$ as the $i$th sample and its label, respectively. Let's define $\bm{z}_i$ as the representation of the $i$th sample in a batch $A$, and $sim(\bm{z}_i, \bm{z}_j) = \frac{\bm{z}_i \cdot \bm{z}_j}{|\bm{z}_i||\bm{z}_j|}$ the cosine similarity, then the SupCon loss~\cite{supcon} is defined as:
\begin{equation}
  \mathcal  L_{\mathrm{SupCon}} =  \sum_{i\in A} \frac{-1}{\left | P(i) \right |} \sum_{p\in P(i)} \log \frac{\exp(sim(\bm{z}_{i}, \bm{z}_{p})/\tau)} {\sum \limits_{a \in A \backslash i} \exp(sim(\bm{z}_i, \bm{z}_a)/\tau)}, 
\label{eq:SupCon}
\end{equation}
where $P(i)$ is the set of positive samples, i.e., with the same label as $z_i$, in the batch excluding $i$ and $a \in A \setminus i$ is the index of all samples in the batch excluding the $i$th sample. $\tau$ is a scalar temperature parameter used for scaling similarity values. The positive pairs are grouped into the numerator and minimizing the loss minimizes their distance in the learned representation and vice versa for negative pairs. SupCon is known to be effective for classifying samples with a single label. However, it is not straightforward to extend this for the multilabel setup as beyond positive samples, where all labels are the same, and negative samples, where none of the labels is the same, there can be a third scenario where labels are partially overlapping. Though SupCon has been extended to hierarchical classification~\cite{himulcon}, it cannot be directly extended to the true multi-label case.

To address this issue, we learn class prototypes as the representative for each class and consider these as anchors while determining positive and negative samples. Specifically, for a specific class prototype, a representation is learned to minimize distances between the prototype and samples with this class label and maximize the distances between the prototype and samples without this class label. We compare the proposed approach with the standard single-label contrastive learning in figure~\ref{fig:lcloss}. We iteratively update the class prototypes while learning the feature representations. We define $C = \{ c_1, \dots, c_K \}$ as the set of classes where $K$ is the number of classes. For a sample $\bm{x}$, let's define \mbox{$P_{ml}(\bm{x}) = \{ c^+_k\}, c^+_k \in C$} as the set of multiple class labels associated with $\bm{x}$ (positive classes) and $c^-_k \in C \setminus P_{ml}(\bm{x})$ denotes the missing classes (negative classes). We define $CP = \{ \bm{cp}_1, \dots, \bm{cp}_K \}$ as the set of class prototypes. Considering $\bm{z}$ is the representation for the sample $\bm{x}$, the class prototypes based multilabel contrastive loss is defined as:
\begin{equation}
\begin{aligned}
    &\mathcal L_{mlc}(\bm{x}) = 
    \\
    &\frac{-1}{|P_{ml}(\bm{x})|} \underset{c^+_k \in P_{ml}(\bm{x})}{\sum}  \! \Big[\log\frac{\exp (sim(\bm{z}, \bm{cp}_k) / \tau)}{\underset{c^-_j \in C \setminus P_{ml}(\bm{x})}{\sum} \exp(sim(\bm{z}, \bm{cp}_j) / \tau)}\Big].
\end{aligned}
\label{eq:loss}
\end{equation}
Here, minimizing the loss of the positive class prototype and instance pairs in the numerator  minimizes the distance between the representation $z$ and the class prototypes corresponding to the sample, and  vice versa for negative classes. We also utilize negative sampling to account for the class imbalance between positives and negatives. 


\vspace{-1mm}
\subsection{Learning class prototypes}
\vspace{-2mm}
\label{sec:prototypes}



We aim to learn class-specific prototypes such that the multilabel samples can be thought of as the combinations of the class prototypes selected based on the associated labels. Lets assume $Z_t$ is an $N \times d$ matrix of $d$-dimensional representations , i.e., $\bm{z}$s), of $N$ samples and $L \in \{ 0, 1\}^{N \times K}$ is corresponding labels matrix with $K$ classes. Let's denote $CP_t$ is a matrix of size $K \times d$ of $K$ class prototypes at a training iteration $t$. Then, \mbox{$Z_t = L \times CP_t + \varepsilon$}, where $\varepsilon$ is the residual noise term. Assuming a Gaussian noise that is unbiased and uncorrelated with the labels $L$, the class prototypes can be approximated as \mbox{$CP^*_t \approx (L^T L)^{-1}L^TZ_t$}, where operation $(L^T L)$ results in a square matrix amenable to inversion. Note that for single labels, this implies averaging the features of the instances belonging to a given class as the prototype for that class. In a multi-label setup, additionally, the co-occurrence between the labels is considered. Finally, the class prototypes are updated with learning iterations as: 
\vspace{-2mm}
\begin{align}
    CP_{t+1} = \beta \cdot CP_t + (1-\beta) \cdot CP^*_t,
    \label{eq:protoupdate}
\end{align}
\noindent where $\beta$ is the decay parameter for the exponential moving averaging. The moving averaging avoids collapsing prototypes across the training iterations~\cite{PCL2021}. 
\vspace{1mm}

\begin{figure}[t]
    \centering
    \includegraphics[width=0.90\linewidth]{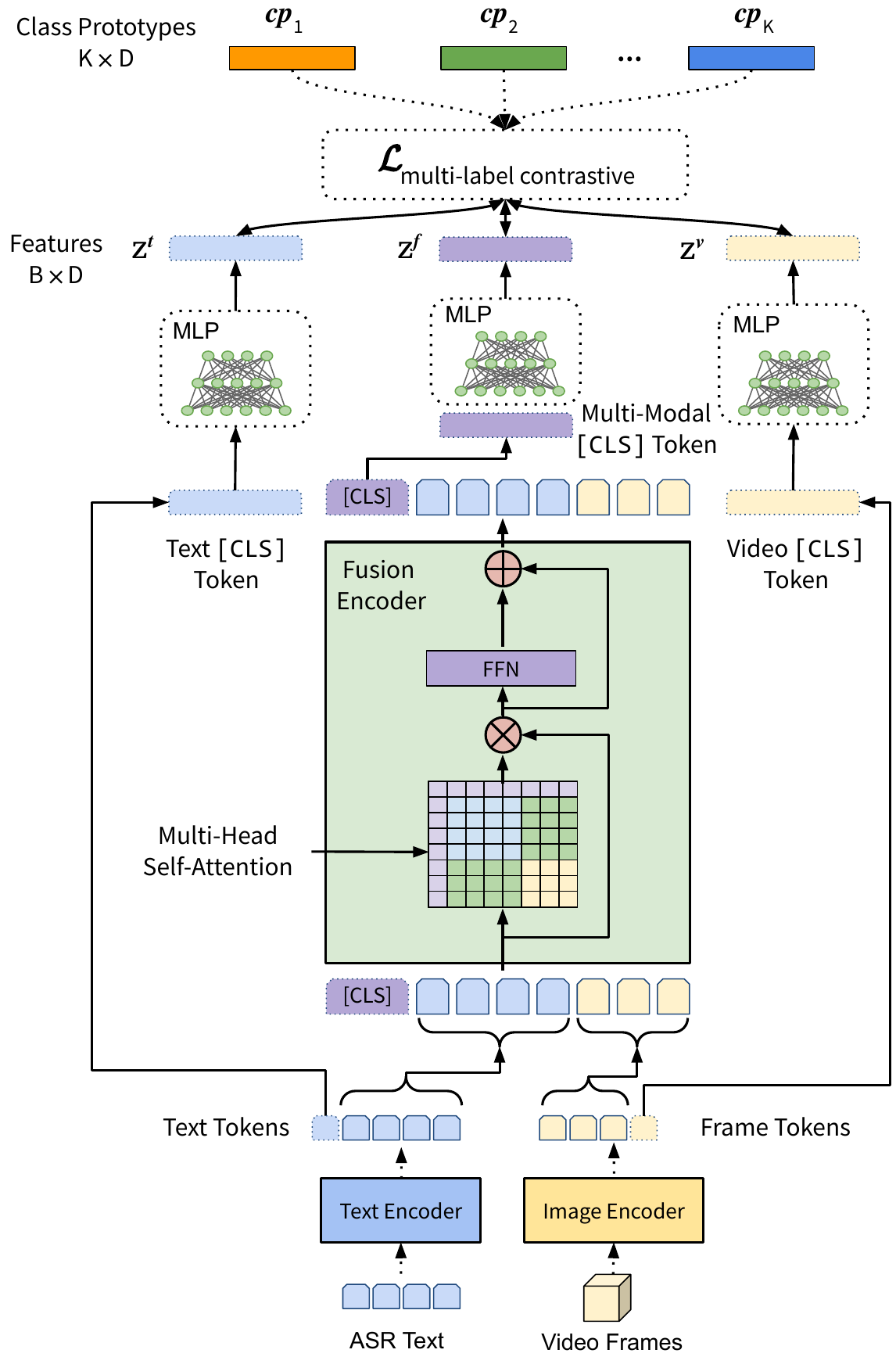}
    \vspace{-2mm}
    \caption{Multi-Modal Classification Network. A text encoder is used to encode  ASR text from the video, while an Image Encoder is used to get tokens representing each frame of the video. Unimodal pre-training is carried out on the text \& image encoders respectively.  Multi-label contrastive loss is used  along with shared prototypes to align the representations across both modalities. This is followed by joint end-to-end learning of the whole multi-modal network including the fusion encoder which applies multi-head self-attention within and across the modalities. The prototypes are further refined during the multi-modal training phase.}
    \label{fig:multitran}
    \vspace{-6mm}
\end{figure}

\vspace{-2mm}
\subsection{Multi-Modal Fusion Transformer}
\vspace{-2mm}
Given a sample video $\bm{x}$, we consider multimodal cues such as visual and audio cues to learn its representation $\bm{z}$. Considering multimodal cues is crucial for content recognition, specifically for education videos as effective comprehension requires attending to both visual demonstration and audio explaining the educational content \cite{national2010common,porter2011common}. To capture audio cues, we consider the audio track of the video and extract speech by removing the background such as tunes or instruments \cite{spleeter2020}. Then we employ the automatic speech recognition (ASR) technique to transcribe text from the speech \cite{radford2022robust}. We notice that separating speech from the background is important for an accurate ASR transcription. Given the video frames and text transcription, we propose a multimodal transformer network (MTN) to fuse these cues using cross-modal attention.

Our MTN (Figure~\ref{fig:multitran}) has three components: image encoder, text encoder, and fusion encoder to learn visual ($\bm{z}^v$), text ($\bm{z}^t$), and fusion ($\bm{z}^f$) representations, respectively. The sample representation is comprised of these three representations $\bm{z} = \{ \bm{z}^v, \bm{z}^t, \bm{z}^f \}$.The image encoder is implemented by a vision transformer (ViT) \cite{dosovitskiy2021image} that learns frame embeddings from the video frames along with a special $CLS$ for each frame \cite{devlin2018bert}. We pool the $CLS$ tokens across the frames and consider this as a compact video representation. Similarly, we consider the BERT-based text transformer \cite{jaegle2021perceiver} to learn the word embeddings along with the text $CLS$ token for the entire transcription. We consider the text $CLS$ as the text representation. Finally, the fusion encoder fuses the visual and text cues by leveraging cross-modal attention between frame and word embeddings.



\vspace{-2mm}
\subsection{Inference using prototypes}
\vspace{-2mm}
Existing approaches in contrastive learning~\cite{supcon, simclr} usually discard the MLP layer after the contrastive training, and a linear classifier is trained on the frozen backbone network. We rely on the class prototypes to carry out inference by utilizing the cosine distance between the learned prototypes and test features. Given our prototype loss-based training, the estimated probability of a given class should be proportional to the normalized temperature scaled cosine distance (Equation~\ref{eq:protoinf}). In practice, we normalize the cosine distance such that $-1$ and $1$ correspond to a confidence of $0$ \& $1$ respectively. Thus the prediction is made following:
\vspace{-1mm}
\begin{align}
    \hat{p}(k|\bm{x}) \propto \exp (sim(\bm{z}, \bm{cp}_k)/\tau),
    \label{eq:protoinf}
\end{align}
where $\bm{z}$ is the multimodal representation of the sample $\bm{x}$.

 \vspace{-2mm}
\subsection{Implementation Details}
\vspace{-1mm}

\noindent \textbf{Training strategy for the multimodal fusion:} We follow a two-stage training process: during the initial unimodal training phase, we utilize \textit{fixed} prototypes in each modality to align the representations. Then in the second stage, we train the unimodal encoders and the multi-modal fusion encoder end-to-end. The cross-modal alignment learned during the first stage improves the learning of the multi-modal representation. The multi-modal learning phase consists of alternating optimization steps of training the network using our contrastive loss per Sec.~\ref{sec:loss} and refining the class prototypes as described in Sec.~\ref{sec:prototypes}.

\noindent \textbf{Image encoder:} \noindent For video frames, \texttt{Random Resized Crop} and \texttt{RandAugment}~\cite{cubuk2020randaugment} augmentations are used from \texttt{torchvision}. We use ImageNet pretrained vision encoders ResNet50~\cite{he2016deep}, ViT-B/32 (224$\times$224 resolution) and ViT-B/16 (384$\times$384 resolution)~\cite{dosovitskiy2021an}.

\noindent \textbf{Text encoder:} We generate text from the videos using Whisper~\cite{radford2022robust}, an open source ASR model. For data augmentation, we generate four versions of the ASR text by back-translation using the \texttt{Helsinki-NLP/opus-mt-\{en-de, en-nl, en-fr\}} models through the \texttt{nlpaug} library~\cite{ma2019nlpaug}. Synonym replacement, text span removal and random word swapping augmentations are also used for the text data.  We use DistilBERT-Base-uncased~\cite{sanh2019distilbert}, and t5-small~\cite{raffel2020exploring} from HuggingFace \texttt{transformers} library as the text encoder.

\noindent \textbf{Optimizer:} \noindent We employ AdamW optimizer~\cite{loshchilov2018decoupled} for training with a learning rate of $0.0005$. Weight decay of 1e-6 is utilized only on the MLP head during contrastive training and the classifier during BCE/Focal/Asym. loss. Since pre-trained vision and text backbones are used, the backbone learning rate is set to 1/10th of the learning rate for the head. Exponential Moving Averaging every 10 steps with a decay of 0.999 was used for the model parameters.

\vspace{-2mm}
\section{Experiments}
\label{sec:experimets}
\vspace{-2mm}

\noindent \textbf{Datasets.} In addition to the proposed APPROVE dataset, we evaluate our approach on a subset of Youtube-8M~\cite{abu2016youtube} and COIN~\cite{coin} datasets.  YT-8M consists of a diverse set of YouTube videos with video and audio modalities. We consider a subset of YT-8M dataset with 46K videos and 165 classes. COIN consists of instructional videos covering a wide variety of domains and spanning over 180 classes.

\noindent \textbf{Baselines.} We compare the efficacy of our multilabel classification framework against the following baselines:

\textbf{Binary cross-entropy:} In this baseline, loss for multiple labels is computed by combining the binary cross-entropy losses for individual classes.

\textbf{Focal loss~\cite{lin2017focal}:} This considers a modified binary cross-entropy to assign a higher weight to hard samples by adjusting a focusing parameter $\gamma$. Negative samples can also be down-weighted by using a weight $\alpha$.  The focal loss for a positive label is given as \mbox{$\mathcal L_{focal}(p) = -\alpha(1 - p)^\gamma log(p)$}. We set $\gamma = 2$ and $\alpha = 0.2$ in our experiments.

\textbf{Asymmetric loss~\cite{ridnik2021asymmetric}:} This builds upon the focal loss by utilizing different focusing parameters $\gamma_+$ and $\gamma_-$ for positive and negative labels. It also ignores the negative samples with a prediction probability lower than a margin $m$. Asymmetric Loss for prediction $p$ corresponding to label $y$ is given as: \mbox{$\mathcal{L}_{asym}(p, y)= -yL_+ - (1-y)L_-$}, where \mbox{$L_+ = (1-p)^{\gamma_+}\cdot\log(p)$} and \\
\mbox{$L_- = (\max(p-m,0))^{\gamma_-}\cdot\log(1-\max(p-m,0))$}.
We follow the 5-step procedure recommended by the original authors to train this baseline. We experimentally set  $\{\gamma_-=2,  \gamma_+=1,  m=0.1\}$ corresponding to the best performance on APPROVE.

\noindent  \textbf{Metrics.} In order to develop a reliable education content detection framework, achieving high precision is crucial. Thus we consider Recall @ 80\% Precision (R@80) as the primary metric. We also consider the standard area under the precision-recall curve (AUPR) that is not sensitive to a specific threshold for making the final prediction. We also consider a label ranking average precision (LRAP)~\cite{Tsoumakas2010} metric that is more suitable for the multilabel setup. This estimates whether the ground truth classes are predicted with higher scores than the rest:
\vspace{-2mm}
\begin{equation*}
    \text{LRAP} = \frac{1}{n}\sum_{i=1}^m \frac{1}{|Y_i|}\sum_{\lambda \in Y_i} \frac{|\lambda' \in Y_i : rnk_i(\lambda') \le rnk_i(\lambda)|}{rnk_i(\lambda)},
\end{equation*}
\noindent where $rnk_i(\lambda)$ is the predicted rank of class $\lambda$ for sample $i$. LRAP is a ranking metric and in independent of a threshold.

\begin{table}[ht]
\centering
\arrayrulecolor{black}
\begin{tabular}{lcclll}
\toprule
\textbf{Subset} & \textbf{Modality} & \textbf{Method} & \textbf{AUPR} & \textbf{LRAP} & \textbf{R@80} \\ 
\cmidrule(r){1-6}
\multirow{10}{*}{\textbf{All}} & \multirow{3}{*}{V} 
                     & BCE     & 45.5 & 54.3 & 6.9 \\ 
                     & & Focal & 45.9 & 56.6 & 15.0 \\ 
                     & & \textbf{Ours}  & \textbf{46.7} & \textbf{57.9} & \textbf{19.6} \\ 
\cmidrule(r){3-6}
                     & \multirow{3}{*}{T} 
                        & BCE   & 79.8  & 85.1 & 63.3 \\ 
                     &  & Focal & 79.9 & 85.7 & 72.8 \\ 
                     &  & \textbf{Ours}  & \textbf{82.5} & \textbf{87.4} & \textbf{75.4} \\
\cmidrule(r){3-6}
                     & \multirow{4}{*}{V+T} 
                       & BCE   & 84.3 & 88.4 & 76.3 \\ 
                     & & Focal~\cite{lin2017focal} & 86.1 & 89.1 & 82.2 \\ 
                     & & Asym.~\cite{ridnik2021asymmetric} & 86.0 & 89.2 & 82.4 \\ 
                     & & \textbf{Ours}  & \textbf{88.4~\improve{+2.3}}  & \textbf{90.7~\improve{+1.5}}  & \textbf{85.5~\improve{+3.1}}  \\
\cmidrule(r){1-6} 
\multirow{3}{*}{\textbf{MTH}} & \multirow{3}{*}{V+T} & BCE          & 86.3          & 92.4 & 80.3 \\ 
                                                 & & Focal          & 87.2          & 92.1 & 82.4 \\ 
                                                 & & \textbf{Ours}  & \textbf{88.4~\improve{+1.2}} & \textbf{93.2~\improve{+1.1}} & \textbf{83.2~\improve{+0.8}} \\
\cmidrule(r){1-6}
\multirow{3}{*}{\textbf{LIT}} & \multirow{3}{*}{V+T} & BCE   & 72.1                   & 82.9 & 50.7 \\ 
                                                 &   & Focal & 72.7                   & 83.5 & 50.9 \\ 
                                                 &   & \textbf{Ours}  & \textbf{73.6~\improve{+0.9}} & \textbf{84.7~\improve{+1.2}} & \textbf{54.7~\improve{+3.8}} \\
\bottomrule
\end{tabular}
\arrayrulecolor{black}
\vspace{-2mm}
\caption{Results on APPROVE dataset. All metrics in \%.\\ V$\rightarrow$Video \& T$\rightarrow$Text. M$\rightarrow$ Math \& L$\rightarrow$ Literacy Subsets.}
\vspace{-1em}
\label{tab:approve}
\end{table}

\begin{table}[ht]
\centering
\arrayrulecolor{black}
\begin{tabular}{cclll}
\toprule
\textbf{Modality} & \textbf{Method} & \textbf{AUPR} & \textbf{LRAP} & \textbf{R@80} \\ 
\cmidrule(r){1-5}
V+T & BCE   & 64.6 & 70.2 & 42.3 \\ 
V+T & Focal~\cite{lin2017focal} & 69.7 & 72.7 & 44.6 \\ 
V+T & \textbf{Ours}  & \textbf{70.9~\improve{+1.2}} & \textbf{74.9~\improve{+2.2}} & \textbf{49.1~\improve{+4.5}} \\
\bottomrule
\end{tabular}
\arrayrulecolor{black}
\vspace{-2mm}
\caption{Results on YT-46K. V$\rightarrow$Video Frames and T$\rightarrow$Text.}
\label{tab:yt46k}
\vspace{-1em}
\end{table}

\begin{table}[ht]
\centering
\arrayrulecolor{black}
\begin{tabular}{ccc}
\toprule
\textbf{Modality} & \textbf{Method} & \textbf{Top-1 Accuracy} \\ 
\cmidrule(r){1-3}
V+T & CE     & 53.7 \\ 
V+T & BCE    & 54.9 \\ 
V+T & Focal~\cite{lin2017focal}  & 56.1 \\ 
V+T & SupCon~\cite{supcon} & 54.7 \\ 
V+T & \textbf{Ours}   & \textbf{57.5}~\improve{\textbf{+1.4}}   \\
\bottomrule
\end{tabular}
\vspace{-2mm}
\arrayrulecolor{black}
\caption{Results on COIN. V$\rightarrow$Video Frames and T$\rightarrow$Text.}
\vspace{-6mm}
\label{tab:coin}
\end{table}

\begin{table*}

	\begin{subtable}[t]{.24\linewidth}
		\tablestyle{4pt}{1.1}
		{		\caption{\textbf{Class Prototypes}} \label{tab:ablation-prototypes}
			\begin{tabular}[t]{@{}l|cc@{}}
				 Variant & APPROVE & COIN  \\
				\hline
				Random       & 84.1 &  56.6 \\
				Orthogonal   & 84.8 &  57.0  \\
				\underline{Learned}      & 85.5 & 57.5  \\
				Hierarchical &  86.0    & 57.8  \\
			\end{tabular}
			
		}		
	\end{subtable}
	\begin{subtable}[t]{.24\linewidth}
		\tablestyle{4pt}{1.1}
		{		\caption{\textbf{Fusion Encoder Size}} \label{tab:ablation-fuse}
			\begin{tabular}[t]{@{}c|cc@{}}
				Layers & APPROVE & COIN  \\
				\hline
				1 & 84.9 &  57.1 \\
				\underline{ 2 } & 85.5 &  57.5  \\
				4 & 85.3 &  57.6 \\
			\end{tabular}
			
		}		
	\end{subtable}
	\begin{subtable}[t]{.24\linewidth}\centering
		\tablestyle{4pt}{1.1}
		{		\caption{\textbf{Vision Encoder}} \label{tab:ablation-vis}
			\begin{tabular}[t]{@{}l|cc@{}}
				Vision Model & APPROVE & COIN  \\
				\hline
				R50      & 84.8 & 55.2  \\
				\underline{ViT-B/32} & 85.5 & 57.5  \\
				ViT-B/16 & 83.8 & 57.8  \\
			\end{tabular}
			
		}		
	\end{subtable}
	\begin{subtable}[t]{.24\linewidth}
		\tablestyle{4pt}{1.1}
		{		\caption{\textbf{Text Encoder}} \label{tab:ablation-lang}
			\begin{tabular}[t]{@{}l|cc@{}}
				Text Model & APPROVE & COIN \\
				\hline
				\underline{DistilBert-B} & 85.5 & 57.5 \\
				t5-S         & 87.3 & 57.9 \\
			\end{tabular}
			
		}		
	\end{subtable}
	\vspace{-3mm}	
	\caption{Ablation studies. R@80\% Precision for APPROVE and Top-1 Accuracy for COIN. Default setup is \underline{underlined}.}\label{tab:1}
	\vspace{-3mm}
	\label{tab:ablation}
\end{table*}

\begin{table*}

	\begin{subtable}[t]{.34\linewidth}
		\tablestyle{4pt}{1.1}
		{		\caption{\textbf{Noisy modalities.}} \label{tab:robust-modality}
			\begin{tabular}[t]{@{}l|cc@{}}
				 \% missing & APPROVE & COIN  \\
				\hline
				0\%    & 85.5 & 57.5   \\
				10\% V & 80.1 & 53.3    \\
				10\% T & 75.8 & 57.2   \\
				30\% T & 68.9 & 42.8   \\
			\end{tabular}
			
		}		
	\end{subtable}
	\begin{subtable}[t]{.3\linewidth}
		\tablestyle{4pt}{1.1}
		{		\caption{\textbf{Run-to-Run Variance}} \label{tab:robust-init}
			\begin{tabular}[t]{@{}l|cc@{}}
				Method & APPROVE & COIN  \\
				\hline
				BCE    & 76.3~$\pm$~0.7 & 54.9~$\pm$~0.6 \\
				Focal  & 82.2~$\pm$~0.5 & 56.1~$\pm$~0.3  \\
				Ours   & 85.5~$\pm$~0.5 & 57.5~$\pm$~0.8 \\
			\end{tabular}
			
		}		
	\end{subtable}
	\begin{subtable}[t]{.35\linewidth}\centering
		\tablestyle{4pt}{1.1}
		{		\caption{\textbf{Initialization}} \label{tab:robust-vis}
			\begin{tabular}[t]{@{}l|cc@{}}
				Method & APPROVE & COIN  \\
				\hline
				 \begin{tabular}[x]{@{}c@{}}ImageNet \& \\Wiki-en+TBC\end{tabular}  & 85.5 & 57.5   \\
     \cmidrule{1-3}
				CLIP      & 86.7 & 63.5 \\
			\end{tabular}
			
		}		
	\end{subtable}
			
	\vspace{-2mm}	
	\caption{Robustness analysis. Our method is robust to partially missing modality and has similar run-to-run variance as baselines.}
	\vspace{-5mm}
	\label{tab:robust}
\end{table*}

\noindent \textbf{Results on APPROVE.}
We compare the proposed approach with the baselines in Tab.~\ref{tab:approve}. Our approach outperforms the strongest baselines by 3.1\% and 2.3\% with respect to R@80 and AUPR, respectively. We also present results for separate models trained on Math and Literacy subsets of Approve respectively. Results on the Math subset are higher compared to the Literacy subset, which indicates the literacy classes are harder to distinguish mostly due to the high inter-class similarity. The top three hardest classes are \texttt{follow\_words}, \texttt{letters\_in\_words}, and \texttt{sounds\_in\_words} and these are from the literacy set.


\noindent \textbf{Results on YT-46K and COIN.}
Beyond the APPROVE dataset, we also test our approach on two public datasets: YT-46K and COIN. Here we provide the results for the multi-modal models and the results for the single-modality models are in Section D of the Supplementary Material.

Results on YT-46K are provided in Table~\ref{tab:yt46k}. As YT-8M was primarily collected with the intention of visual classification, the additional use of text data leads to a smaller improvement compared to APPROVE. Results on COIN are provided in Table~\ref{tab:coin}. As each video from COIN is mapped to a single task. Thus, we consider the Top-1 accuracy as the metric. On COIN we compare our approach with SupCon~\cite{supcon} which is effective for single labels. Note that our approach outperforms SupCon and this justifies the effectiveness of the class prototypes based training in a generic contrastive learning framework.



\subsection{Ablation Studies}
\vspace{-2mm}
We perform the following ablation studies to quantify the impact of our approach for learning class prototypes, the multimodal fusion module, and the choice of visual and text encoding frameworks.

\noindent \textbf{Learning class prototypes:} We compare the two strategies where after initializing the class prototypes, we 1) keep them fixed and learn only the multimodal embedding of the samples, and 2) class prototypes and sample embedding are learned iteratively. The initialization can be done either randomly, or with orthogonal constraints. We note that orthogonal initialization performs best in our experiments and iterative adjusting the class prototypes achieves better performance as shown in Table~\ref{tab:ablation} (a). We also consider hierarchical prototypes, for APPROVE, using a 2-level hierarchy where the first level consists of 18 classes, and the second level is the 3 super-classes: math, literacy, and background. The 180 task categories of COIN are organized into 12 domains in the taxonomy provided with the dataset. This hierarchy imposes an additional constraint on learning the embeddings during training. 

\noindent \textbf{Fusion Encoder:} This evaluates the effect of the number of layers in the fusion encoder on the final performance. As expected. the performance improves with more layers and saturates around 4 layers (Table~\ref{tab:ablation} (b)). 

\noindent \textbf{Vision Encoder:} This evaluates the effect of the image encoder on the final performance. We consider ResNet \cite{he2016deep} and ViT variants \cite{dosovitskiy2021image}. We notice that the ViT-B/16-384 encoder works well for the larger COIN dataset, whereas the ViT-B/32 encoder works best for APPROVE (Table~\ref{tab:ablation} (c)).

\noindent \textbf{Text Encoder:} This evaluates the effect of the text encoder on the final performance. We test DistilBERT and T5 backbones for the text encoder. BERT is trained to predict masked spans of text. T5's unsupervised objective is similar, however, it trains on predicting the entire sequence instead of just the masked spans. GPT2 takes an auto-regressive approach to language modeling (Table~\ref{tab:ablation} (d)).


\vspace{-2mm}
\subsection{Robustness analysis}
\vspace{-2mm}

\noindent We evaluate the robustness of our approach as follows:

\noindent \textbf{Noisy modality:} YouTube videos may have noisy modalities where some of the video frames are missing or ASR transcription is noisy. We show that our approach is robust against the cases where a percentage of video frames or text words are missing as shown in Tab~\ref{tab:robust}(a).

\noindent \textbf{Run-to-Run variance:} The low variance across runs (Tab~\ref{tab:robust}(b)) indicates that our approach is not sensitive to random initialization of the class prototypes.

\noindent \textbf{Initializing the encoders:} We consider the ImageNet pre-training for the image encoder, while English Wikipedia + Toronto Book Corpus is used to pre-train the text encoder. We provide results where the backbones are initialized with  CLIP~\cite{radford2021learning}, which provides a more aligned vision-text representation. As expected, the results are better with the CLIP initialization (Tab~\ref{tab:robust}(c)). The improvements are more significant on COIN than APPROVE as CLIP models may not be exposed to educational videos. 
\vspace{-2mm}
\section{Conclusion}
\label{sec:conclusion}
\vspace{-2mm}

We have proposed an approach for detecting educational content in online videos. The problem is formulated as a fine-grained multilabel video classification task and we have considered class-prototypes based contrastive learning to address this. We have employed a multimodal transformer network to fuse visual and audio cues. This is crucial for comprehending educational content as both visual and audio cues are to be aligned to ensure effective comprehension. Our approach is shown to be effective in distinguishing fine-grained educational content with high inter-class similarity. We have introduced APPROVE - a dataset with 193 hours of expert-annotated educational videos. Beyond APPROVE, we have evaluated our approach on COIN and YouTube-8M datasets where our approach outperforms the competitive baselines.



\vspace{-3mm}
\section*{Acknowledgements}
\vspace{-2mm}
\noindent {\setstretch{0.9} {\small This work is supported by NSF grant \# 2139219 and SRI R\&D award.~Rohit Gupta is supported by ARO grant W911NF-19-1-0356. Views and conclusions contained herein are those of the authors and should not be interpreted as representing the official policies or endorsements, either expressed or implied, of NSF, ARO, IARPA, DOI/IBC, or the U.S. Government. The U.S. Government is authorized to reproduce and distribute reprints for Governmental purposes notwithstanding any copyright annotation thereon.}\par}

\onecolumn
\section*{Supplementary Material}

\noindent This supplementary material is organized into four sections. In Section~\ref{sec:approve} we provide additional details about our APPROVE dataset. Section~\ref{sec:classwise} presents detailed classwise result for our models, demonstrasting the impact of multi-modal learning. In Section~\ref{sec:features} we analyze the feature space of our trained model to demonstrate that it picks up semantic similarities between the classes. Finally, we provide implementation details in Section~\ref{sec:implement} to assist reproducibility.  Note that this document contains 12 pages and some are partially blank to clearly separate different parts of the material.

\appendix

\section{APPROVE Dataset}
\label{sec:approve}
The key property of APPROVE that sets it apart from prior datasets is its fine-grained and multi-label nature. We provide some visualizations here to build on the main paper and illustrate these properties.

\subsection{Fine-Grained}

Additional samples from the Literacy (in Figure~\ref{fig:litsamples}) and Math (in Figure~\ref{fig:mathsamples}) splits of APPROVE are visualized here. These examples are randomly picked, and they highlight the fine-grained nature of the dataset. Many pairs or groups of classes in APPROVE have very high visual similarity, e.g. \texttt{Shape ID} and \texttt{Building Drawing Shapes}; \texttt{Sounds in Words} and \texttt{Rhyming}, etc. Also note that background, math and literacy are distinct and do not share overlapping labels, which illustrates the heirarchical struture of APPROVE. 

\subsection{Multi-Label}

\begin{figure*}
    \centering
    \includegraphics[width=\linewidth]{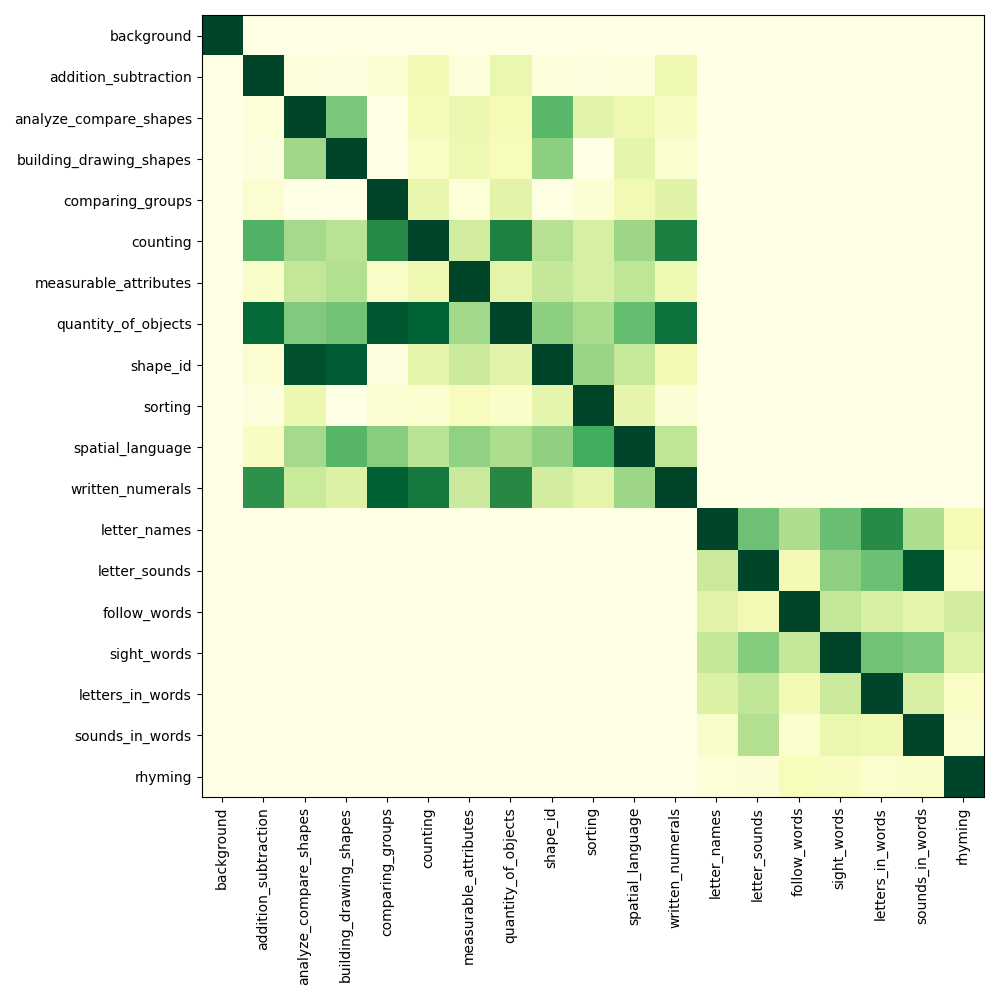}
    \caption{Ground=Truth Multi-Label co-ccurence matrix for APPROVE videos. The three high level groups of categories: Background, Math and Literacy can be seen, highlighting the hierarchical structure of the dataset.}
    \label{fig:coccurence}
\end{figure*}

APPROVE is a densely multi-labelled dataset. The multi-label co-occurence matrix for APPROVE is visualized in Figure~\ref{fig:coccurence}. Each cell of the matrix, $L[i,j]$, equals the fraction of videos with class $i$ which also contain class $j$. Note that the matrix is not symmetric as two classes might have a one-sided relationship. e.g. presence of \texttt{written numerals} suggests \texttt{comparing groups} is highly likely to be present, however the inverse does not hold. since many videos teach how to compare groups without using written numberals, e.g. comparing groups of objects by some non-numeric property such as a shape or color. 

\subsection{Class descriptions}
Detailed description of the education codes corresponding to each class in APPROVE and their annotation criteria are presented in Table~\ref{tab:codes}. These fine-grained classes correspond to age-appropriate curriculum topic recommendations prescribed by the common core education standards. These detailed descriptions along with a large batch of examples was provided to all data annotators to ensure high quality labeling. 

\subsection{Annotation evaluation}
To ensure the quality and correctness of the annotations, we consider educational researchers to annotate the videos and follow a standard validation protocol \cite{Radesky2020}. We consider two expert annotators and a few education researchers for annotating the videos. Experts train annotators to identify the indicators of educational content in videos. After training, education researchers are evaluated on a validation set of 50 videos that are already annotated by two experts. Annotators are allowed to start the final annotation process once they achieve more than 90\% agreements with the expert annotations. We observe a 95\% agreement is reached after four weeks of training. We also consider inter-annotator consistency for the final annotations. Videos that are not consistently labeled by all the annotators are ignored.  

\subsection{Education codes}
\begin{figure*}
    \centering
    \includegraphics[width=\linewidth]{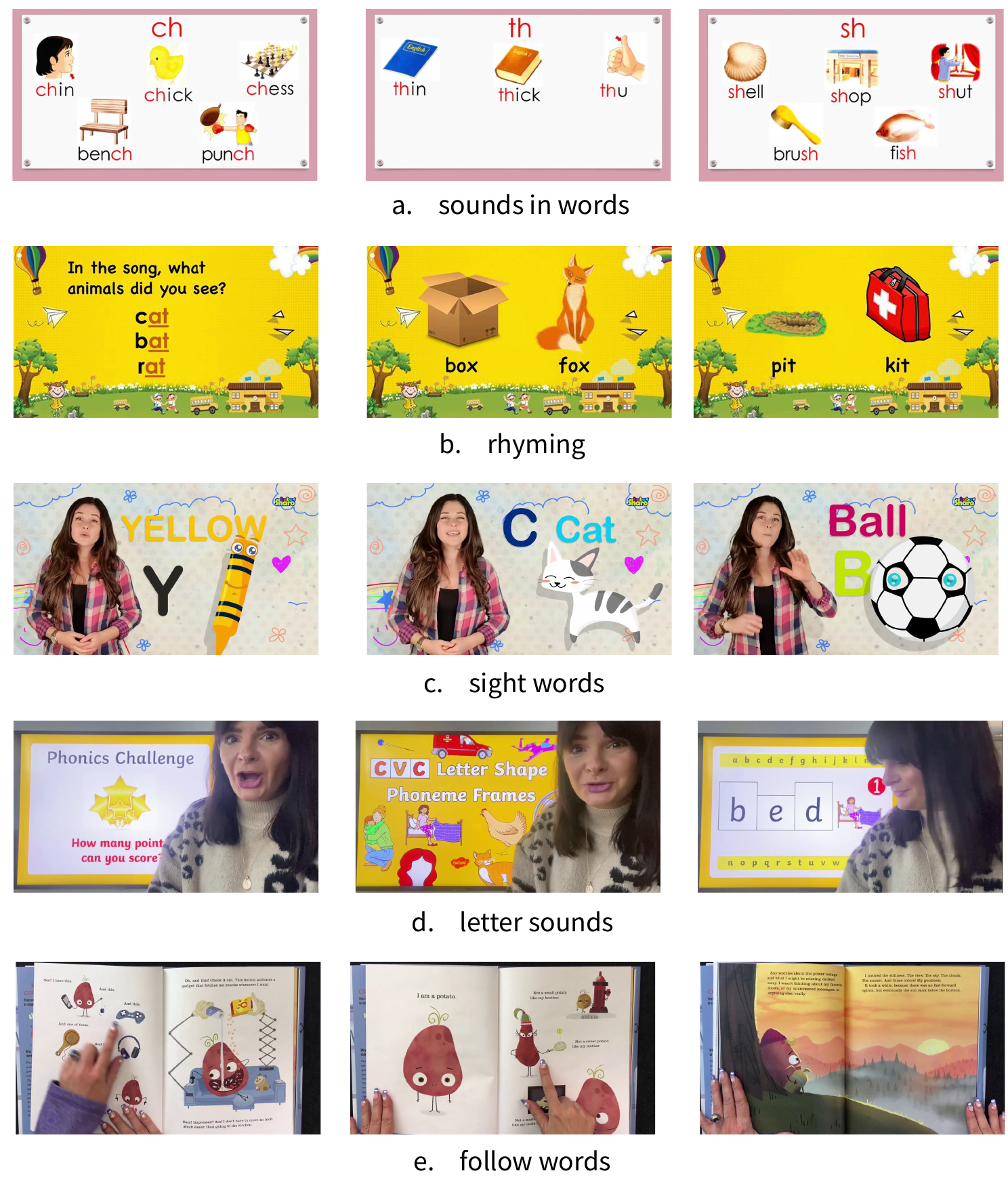}
    \caption{Sample frames from five Literacy classes in APPROVE. The classes share visual similarity, which makes classification a challenging fine-grained learning task.}
    \label{fig:litsamples}
\end{figure*}

\begin{figure*}
    \centering
    \includegraphics[width=\linewidth]{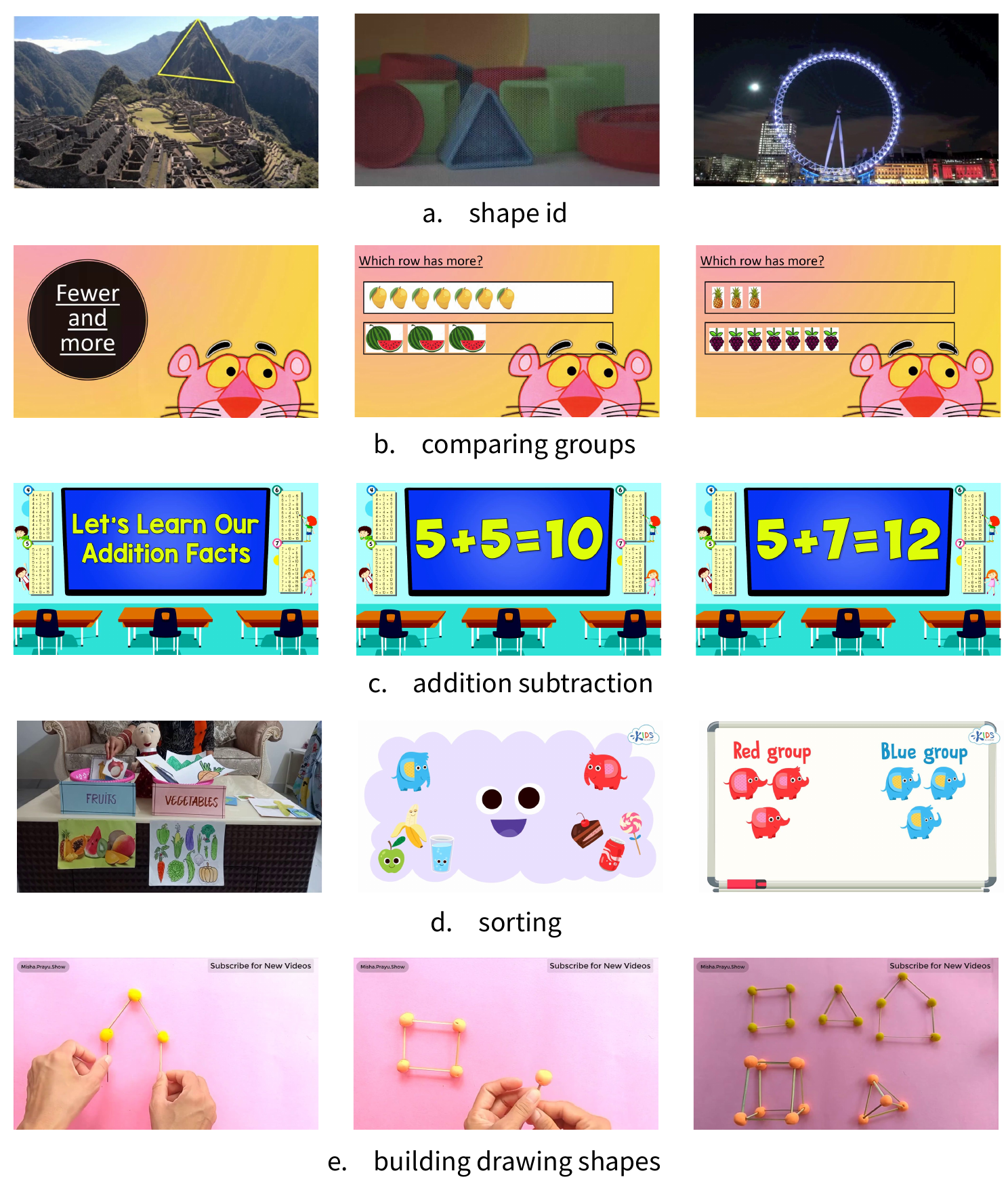}
    \caption{Sample Frames from five Math classes in APPROVE.}
    \label{fig:mathsamples}
\end{figure*}

APPROVE consists of 193 hours of videos with 19 classes including 7 literacy codes, 11 math, and background. We follow the Common Core Standards \cite{national2010common,porter2011common} to select education content suitable for the kids at kindergarten level. Descriptions of these codes are provided in Tab.~\ref{tab:codes}. Some sample frames for these codes are presented in Fig.~\ref{fig:litsamples} and \ref{fig:mathsamples}. 

\clearpage

\begin{small}
\renewcommand*{\arraystretch}{1.1}
\begin{longtable}{m{3cm} |  m{1.5cm}  | m{10.0cm} }
\caption{Description of the education codes used in APPROVE. These codes correspond to age-appropriate curriculumn recommendations prescribed by the common core education standards.} \label{tab:codes} \\

\hline \multicolumn{1}{c|}{\textbf{Code names}} & \multicolumn{1}{|c|}{\textbf{High-level class}} & \multicolumn{1}{|c}{\textbf{Description of the code}} \\ \hline
\endfirsthead

\multicolumn{3}{c}%
{{\bfseries \tablename\ \thetable{} -- continued from previous page}} \\
\hline \multicolumn{1}{c|}{\textbf{Code names}} & \multicolumn{1}{|c|}{\textbf{High-level class}} & \multicolumn{1}{|c}{\textbf{Description of the code}} \\ \hline
 \endhead

\hline \multicolumn{3}{c}{{\textbf{Continued on next page}}} \\ \hline \hline
\endfoot

 \hline
 \multicolumn{3}{c}{\textbf{End of Table 1}}\\
 \hline  \hline
 \endlastfoot

\hline
Counting  & Math &
\begin{itemize}
        \item More than one number in the standard sequence. Starting point can be any whole number. 
        \item This includes counting parts of a shape, such as counting sides or vertices.
    \end{itemize} \\ 
\hline
Written numerals  & Math & 
\begin{itemize}
        \item A written numeral, either on its own or as part of a count sequence, with corresponding visual or audio.
    \end{itemize} \\ 

\hline
Quantity of objects  & Math & 
\begin{itemize}
        \item Emphasizing that the last number in a count sequence represents the total number of objects.
    \end{itemize} \\ 

\hline
Addition or subtraction  & Math & 
\begin{itemize}
        \item At least two numbers and the number that results when they are added or subtracted.
        \item Includes adding by counting on. For example, “We have three, let’s add two. Three, four, five. Three and two make five.”
        \item Includes decomposing sets of objects into two or more sets. 
    \end{itemize} \\ 

\hline
Measurable attributes  & Math & 
\begin{itemize}
        \item Describing one object or comparing multiple objects based on at least one measurable attribute, such as length or volume.
    \end{itemize} \\ 

\hline
Comparing groups  & Math & 
\begin{itemize}
        \item Comparing two or more groups of objects.
    \end{itemize} \\ 

\hline
Sorting  & Math & 
\begin{itemize}
        \item Objects being sorted into categories, such as but not limited to color, shape, object type, purpose, pattern, or species.
    \end{itemize} \\ 

\hline
Spatial language  & Math & 
\begin{itemize}
        \item Words and visuals to describe position or movement.
    \end{itemize} \\ 

\hline
Shape identification  & Math & 
\begin{itemize}
        \item Naming and displaying a shape.
    \end{itemize} \\ 

\hline
Building or drawing shapes  & Math & 
\begin{itemize}
        \item Showing how a geometric shape is drawn or built. 
    \end{itemize} \\ 

\hline
Analyzing or comparing shapes  & Math & 
\begin{itemize}
        \item Describing one or more shapes in terms of their attributes.
    \end{itemize} \\ 

\hline
Letter names & Literacy & 
\begin{itemize}
        \item Any spoken or sung letter name. 
        \item Does not need to say whether the letter is upper- or lowercase.
    \end{itemize} \\ 

\hline
Letter sounds & Literacy & 
\begin{itemize}
        \item Any spoken or sung letter sound. Must be distinct from a spoken word.
        \item Can include letter names if they are also letter sounds (e.g., long forms of vowels).
        \item Does not need to say whether the letter is upper- or lowercase.
    \end{itemize} \\ 

\hline
Letters in words & Literacy &
\begin{itemize}
        \item A letter within a word is visually highlighted.
        \item If the video names multiple letters within a word, to meet this criterion it must highlight each letter individually as it is named. 
        \item If the video separately names the letter and then displays it in a word, the letter must be visually highlighted within the word. 
    \end{itemize} \\ 

\hline
Sight words  & Literacy & 
\begin{itemize}
        \item Only words on the sight words list count for this code. As long as a video includes at least one word on the sight words list, this indicator is present.
    \end{itemize} \\ 

\hline
Sounds in words & Literacy & 
\begin{itemize}
        \item The sound may occur anywhere within a word, including the beginning sound.
        \item The sound must not be the full word. 
        \item Choose response option, sound of individual letter if the video includes audio of the sound a single letter makes within a word.
        \item Choose response option, sound of multiple letters together if the video includes audio of the sound of two or more letters together within a word, excluding the full word. For example, “at” in “rat.”
        \item Can include separately making the sound of a letter on its own and displaying it in a word, so long as both occur within about 2 seconds.  
        \item Still counts even if other words are used between the full word and the sound. For example, “The words cat and rat both have the ‘t’ sound at the end.”
    \end{itemize} \\ 
\hline
Follow words  & Literacy & 
\begin{itemize}
        \item Must show a passage containing multiple words. Only one word on screen at a time would not count. 
        \item Words must be highlighted left to right, top to bottom, and/or page to page. Highlighting can include one word in a passage appearing at a time.
        \item It’s okay if words aren’t highlighted exactly as they are spoken (e.g., highlighting an entire line of text in a paragraph at a time, highlighting words at a constant pace that doesn’t totally line up with audio), so long as the highlighting generally moves left to right or top to bottom as the words are spoken.
        \item Includes sing-along style videos that highlight words as they are sung.
    \end{itemize} \\ 

\hline
Rhyming  & Literacy & 
\begin{itemize}
        \item Within 60 seconds of the word "rhyme" or "rhyming," audio of at least 2 rhyming words. 
        \item “Rhyme” may occur before or after the rhyming words. 
        \item Rhyming words do not need to be spoken one after the other (e.g., “cheese, please”); they could have words between them, such as a poem or song (e.g., the cat jumped over the hat).
    \end{itemize} \\ 

\hline
\end{longtable}
\end{small}
 \clearpage

\section{Classwise Results}
\label{sec:classwise}

\begin{figure*}[h]
    \centering
    \includegraphics[width=0.95\linewidth]{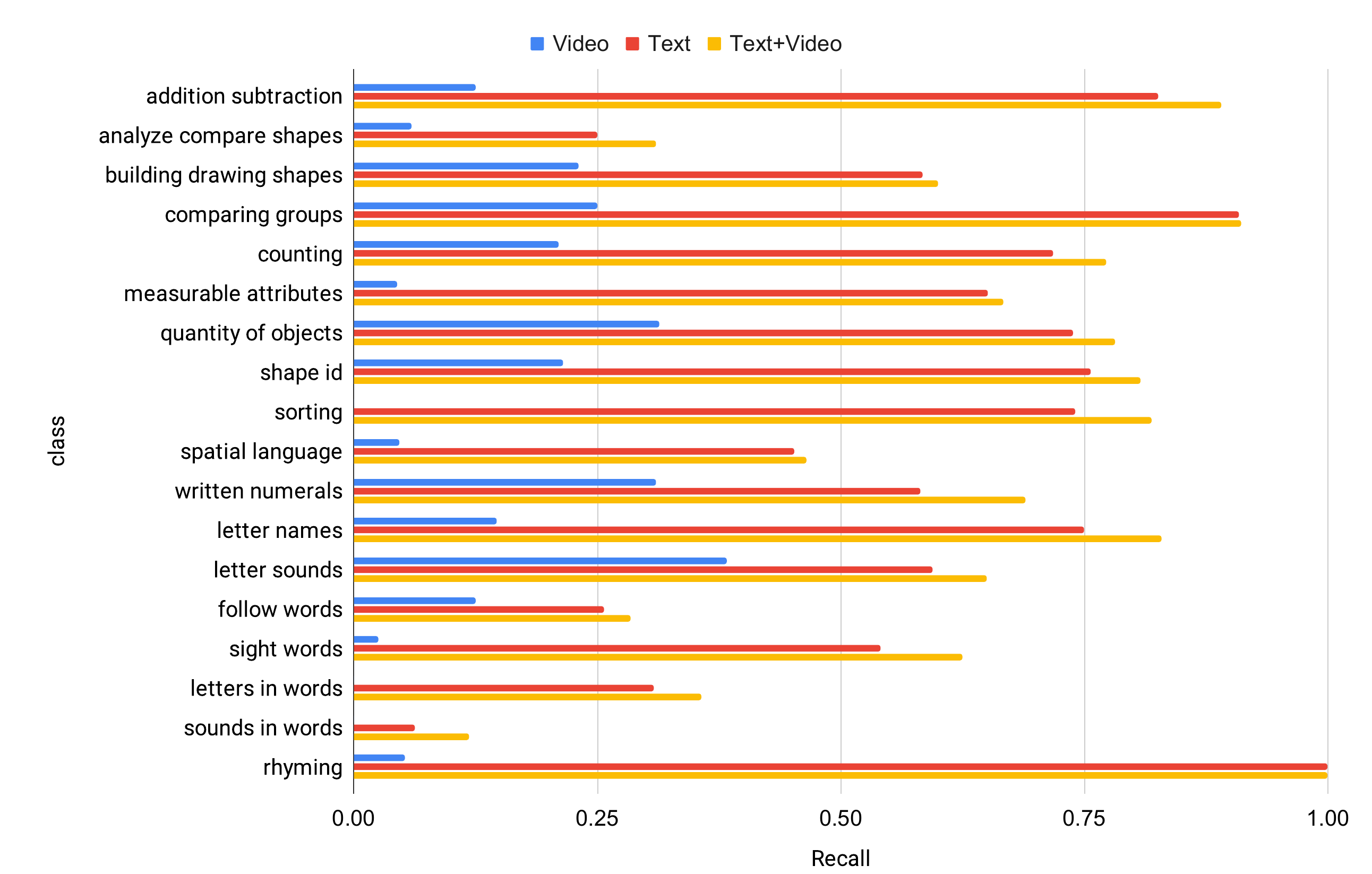}
    \vspace{-3mm}
    \caption{Classwise Recall at 80\% overall Precision. Most classes benefit from access to multi-modal input data.}
    \label{fig:recall}
\end{figure*}

\begin{figure*}[h]
    \centering
    \includegraphics[width=0.95\linewidth]{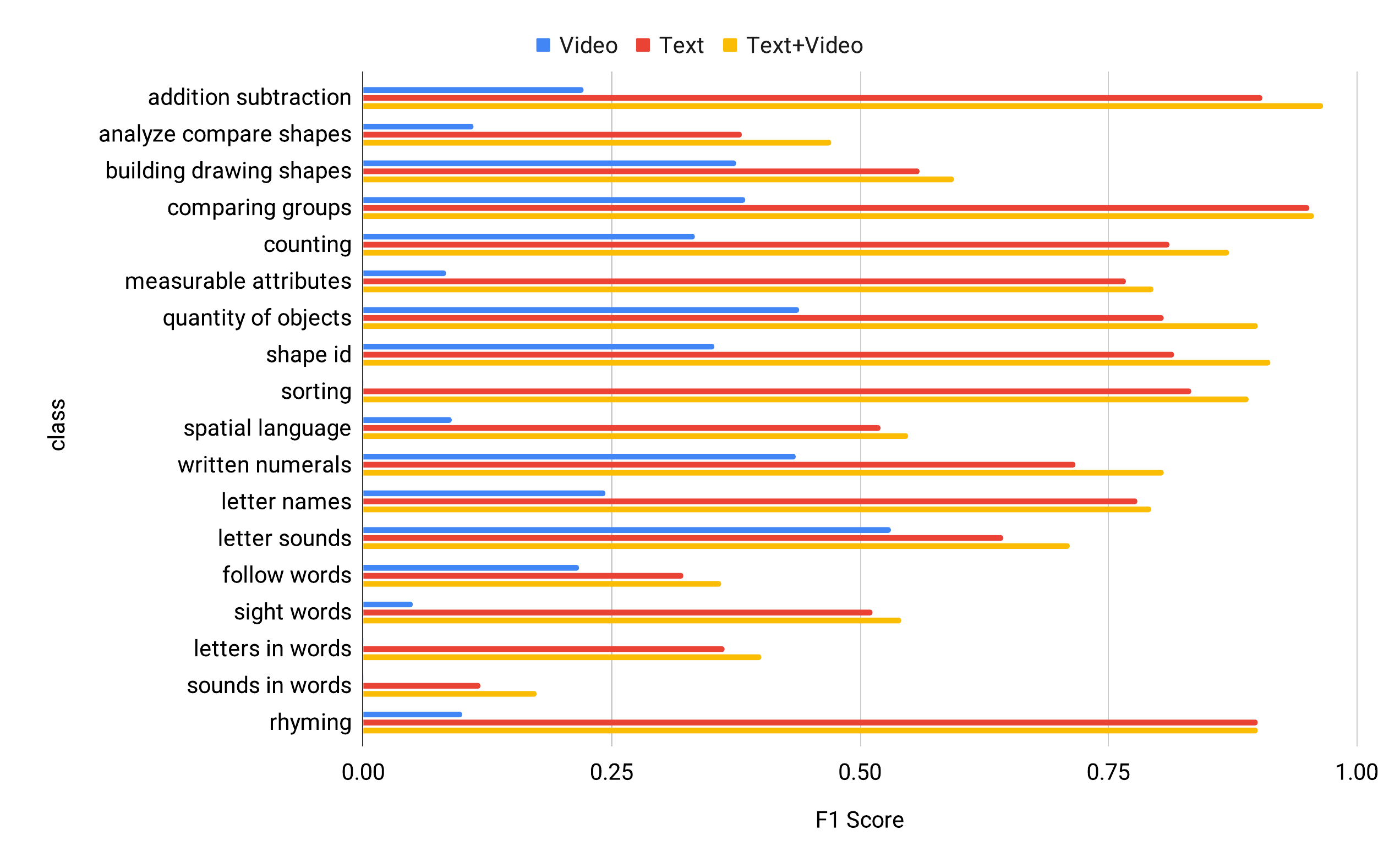}
    \vspace{-5mm}
    \caption{Classwise F1 Scores at 80\% overall Precision. }
    \label{fig:f1s}
\end{figure*}

We demonstrated strong overall results in the main paper. In particular we found that using multi-modal input data resulted in strong results. Here, we provide class-wise recall and F1 scores in Figure~\ref{fig:recall} and Figure~\ref{fig:f1s} respectively. These show that our improvements occur across a wide variety of video classes. In order to compute Recall and F1 score, we pick the classification threshold to achieve 80\% overall precision to satisfy the requirements of the sensitive education application scenarios (as discussed in the main paper). The threshold found are 0.91 for Video only model, 0.56 for the Text only model and 0.51 for the Video+Text model. As can be noticed in Figure~\ref{fig:recall} Text only model generally outperforms the Video only model, but the Text+Video model outperforms the Text only model for most classes. Classes which focus on skills requiring connecting language to vision such as \texttt{Sight Words}, \texttt{Written Numerals} and \texttt{Sorting} benefit the most from the use of multi-modal data for classification. 

\begin{figure*}[h]
    \centering
    \includegraphics[width=\linewidth]{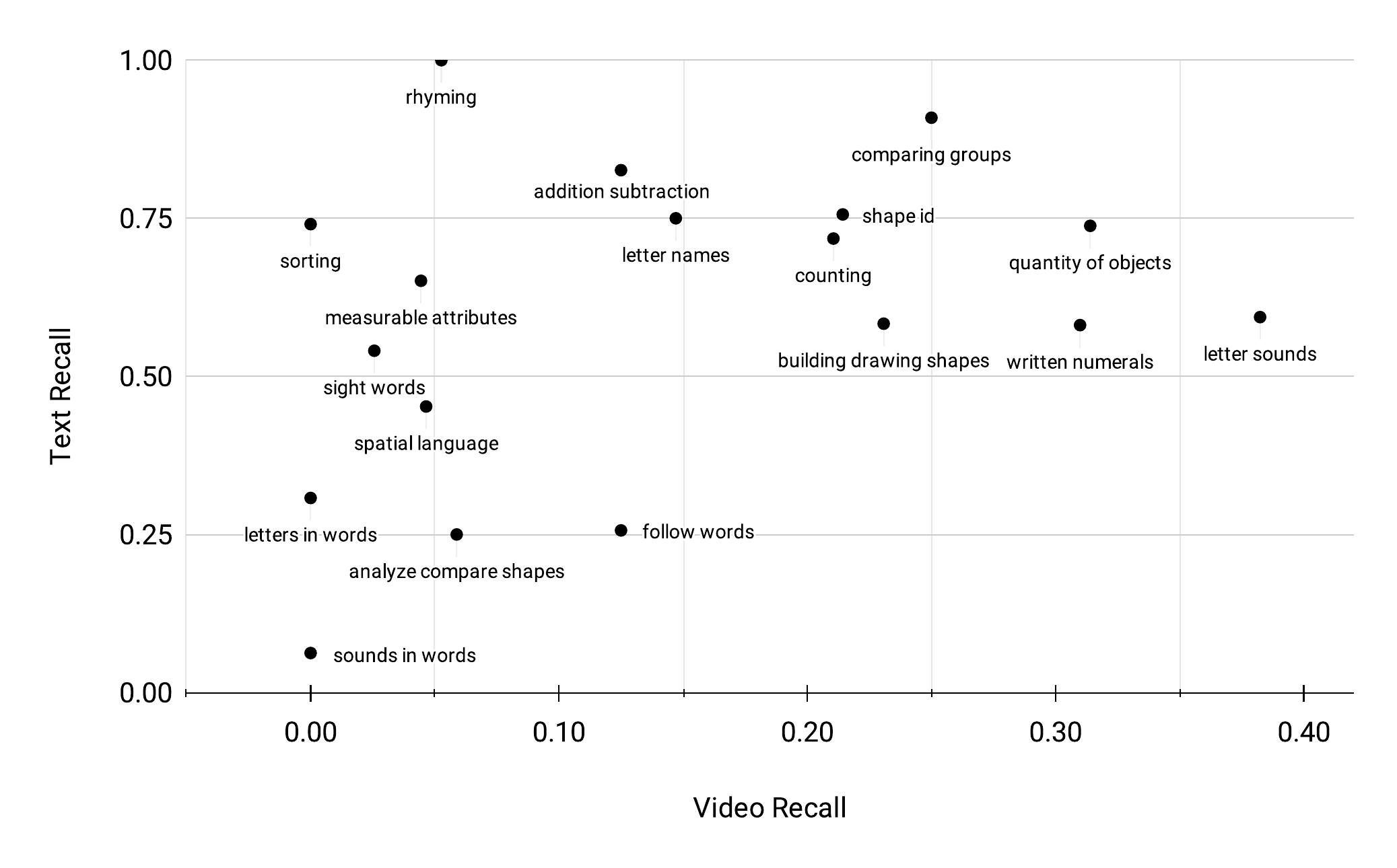}
    \vspace{-5mm}
    \caption{Comparing classwise recall between video and text models.}
    \label{fig:compare}
\end{figure*}

In Figure~\ref{fig:compare} we provide a scatter plot of class-wise recall for the text only model recall vs the video only model. The recall is weakly correlated across the two modalities (R~\textsuperscript{2} = 0.122), which explains the significant gains due to combining the two modalities.

\clearpage

\begin{figure*}[h]
\centering

\begin{subfigure}[b]{0.7\linewidth}
    \centering
    \includegraphics[width=\linewidth]{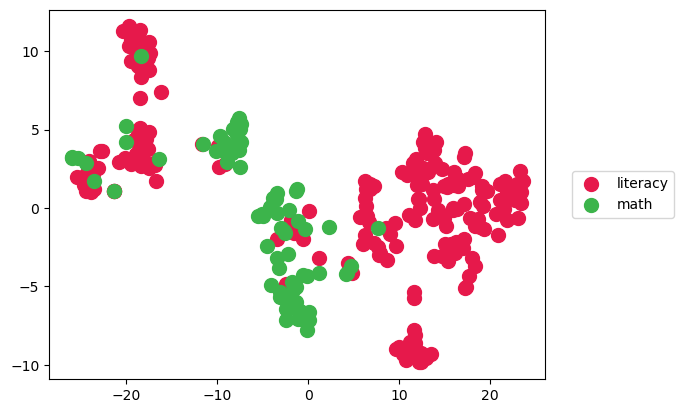}
    \caption{Top Level Classes}
\end{subfigure}
\hfill \\
\begin{subfigure}[b]{0.32\linewidth}
    \centering
    \includegraphics[width=\linewidth]{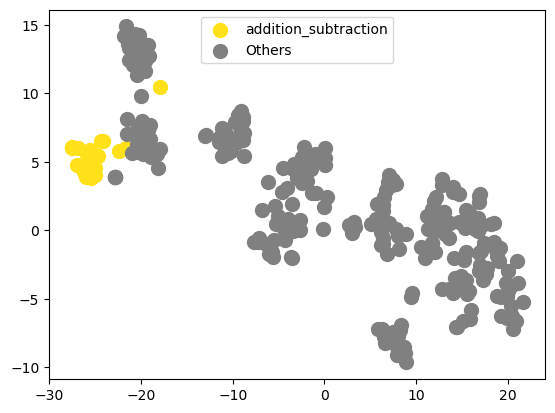}
    \caption{addition subtraction}
\end{subfigure}
\hfill
\begin{subfigure}[b]{0.32\linewidth}
    \centering
    \includegraphics[width=\linewidth]{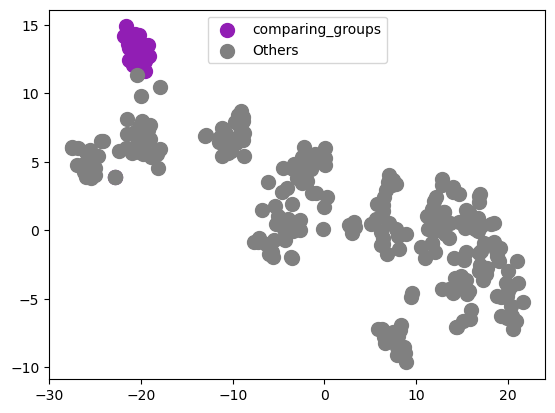}
    \caption{comparing groups}
\end{subfigure}
\hfill
\begin{subfigure}[b]{0.32\linewidth}
    \centering
    \includegraphics[width=\linewidth]{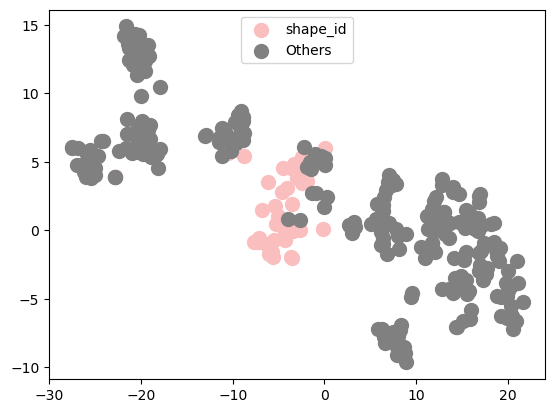}
    \caption{shape id}
\end{subfigure}
\begin{subfigure}[b]{0.32\linewidth}
    \centering
    \includegraphics[width=\linewidth]{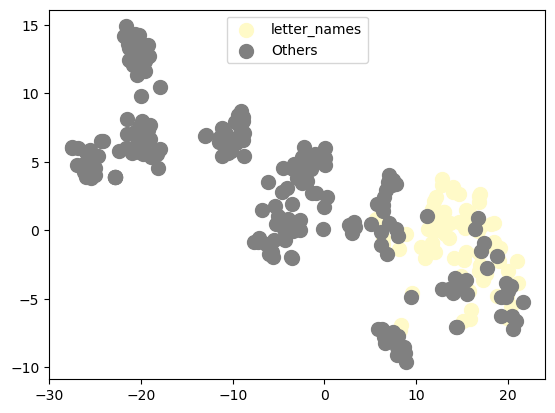}
    \caption{letter names}
\end{subfigure}
\hfill
\begin{subfigure}[b]{0.32\linewidth}
    \centering
    \includegraphics[width=\linewidth]{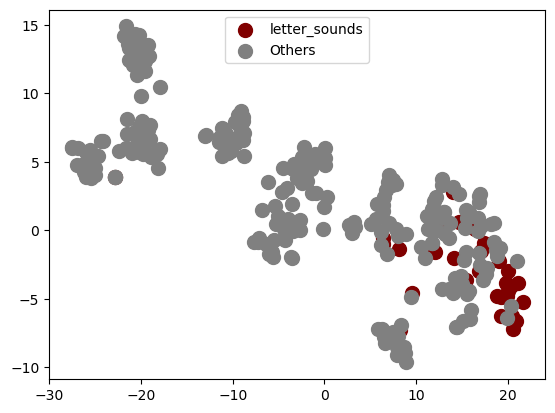}
    \caption{letter sounds}
\end{subfigure}
\hfill
\begin{subfigure}[b]{0.32\linewidth}
    \centering
    \includegraphics[width=\linewidth]{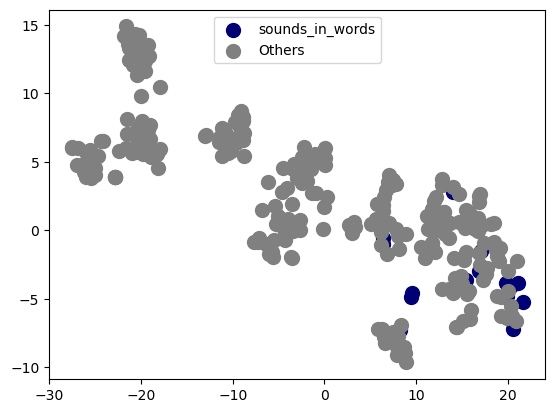}
    \caption{sounds in words}
\end{subfigure}

\caption{Visualizing features for APPROVE test set samples using tsne. \textbf{(a)} Top level classes, math and literacy, are disjoint in feature space. \textbf{(b-g)} Since APPROVE is a multi-label dataset, we show one-vs-all tsne plots.}
\label{fig:tsne}
\end{figure*}

\section{Learned Feature Representation}
\label{sec:features}

We visualize learned features from our model using t-SNE in Figure~\ref{fig:tsne}. At the top level literacy and math videos are cleanly separated. We also plot one-vs-all plots for each class as APPROVE is a multi-label dataset. It can be observed that even fine-grained classes are well clustered, especially for math topics.

\newpage
\section{Implementation Details}
\label{sec:implement}
\subsection{Models}

\noindent \textbf{Image Backbone:} \noindent We use ImageNet pretrained ResNet50~\cite{he2016deep} and Instagram user generated tag weakly supervised (SWAG) + ImageNet finetuned ViT-B/16 (384$\times$384 resolution)~\cite{dosovitskiy2021an} from the TorchVision model zoo. We also use the ImageNet-21K pre-trained and ImageNet finetuned ViT-B/32 (224$\times$224 resolution) from HuggingFace. 

\noindent \textbf{Text Backbone:} For the language encoder, we use DistilBERT-Base-uncased~\cite{sanh2019distilbert}, and t5-small~\cite{raffel2020exploring} from the huggingface \texttt{transformers} library.

\subsection{ASR Generation:} We generate ASR text from the videos using OpenAI-whisper~\cite{radford2022robust}, which is an open source ASR model. We used the \texttt{medium.en} version of the model and turned off the \texttt{condition\_on\_previous\_text} option as we observed that ASR generation would collapse for videos with poor audio quality with that option turned on by default. 

\subsection{Other Datasets}

Some video classification datasets have been proposed with class labels based on \textit{topics} being discussed or illustrated. One such dataset is COIN~\cite{coin}, which consists of instructional videos from 180 diverse coarse-grained tasks covering a wide range from \texttt{changing-car-tire} to \texttt{making-pizza}. While temporal sub-task segmentation labels are also available, in this paper we restrict ourselves to fine grained video classification task. YouTube-8M(YT-8M)~\cite{abu2016youtube} dataset is a large sample of YouTube data labeled with many  coarse-grained visual entities, however, because of its large size, its distributed in the form of extracted visual and audio features. In order to fully test the potential of our method on this dataset, we create a subset using 1\% of YT-8M data called YT-46K, which consists of 46,000 videos (note that despite its name the full YT-8M only contains 5.6 Million videos, since we scraped a 1\% shard, we attempted scraping about 56,000 videos, of which about 46,000 were still available) with full video, audio and text metadata scraped from YouTube. Since it is a long tailed multi-label dataset, the frequency of labels follows a power-law distribution. We restrict the number of classes to those with at least 100 instances, which results in 166 usable labels.


\clearpage

\onecolumn
\subsection{Data Augmentations}

We use \texttt{RandAugment} and \texttt{RandomResizedCrop} for augmenting video frames. RandAugment magnitude is ramped up from 1 to 10 over first 20 epochs.

\begin{lstlisting}[language=Python]
import torchvision.transforms as t

t.Compose([t.RandAugment(magnitude=magnitude),
           t.RandomResizedCrop(224, scale=(0.4, 1.0), \ 
           interpolation=t.InterpolationMode.BILINEAR),
           t.ConvertImageDtype(torch.float32),
           t.Normalize((0.5, 0.5, 0.5), (0.5, 0.5, 0.5)),
           ])
\end{lstlisting}





\noindent For augmenting text datat we use synonym replacement from paraphrase dataset, random span cropping and random word swapping. As Back Translation is computationally expensive, we compute 4 additional back translated versions of each text before training.

\begin{lstlisting}[language=Python]
import nlpaug.augmenter.word as naw
import nlpaug.flow as naf

# m represents the magnitude of augmentation
m = 0.5 
t = [naw.SynonymAug(aug_src="ppdb", aug_min=2, aug_max=15, aug_p=m)]
t += [naw.RandomWordAug(action="crop", aug_min=1, aug_max=5,  aug_p=m)]
t += [naw.RandomWordAug(action="swap", aug_min=1, aug_max=5, aug_p=m/2.)]
train_augmenter = naf.Sequential(t)


from nlpaug.augmenter.word \ 
           .back_translation import BackTranslationAug as BTAug

# actual arguments to BTAug are from_model_name, to_model_name
# abbreviated to fit the command in one line 
BTAug(from='facebook/wmt19-en-de', to='facebook/wmt19-de-en')
BTAug(from='Helsinki-NLP/opus-mt-en-de', to='Helsinki-NLP/opus-mt-de-en')
BTAug(from='Helsinki-NLP/opus-mt-en-nl', to='Helsinki-NLP/opus-mt-nl-en')
BTAug(from='Helsinki-NLP/opus-mt-en-fr', to='Helsinki-NLP/opus-mt-fr-en')
\end{lstlisting}





\newpage

{\small
\bibliographystyle{ieee_fullname}
\bibliography{egbib}
}

\end{document}